\def\year{2019}\relax
\documentclass[letterpaper]{article} 
\usepackage{aaai19}  
\usepackage{times}  
\usepackage{helvet}  
\usepackage{courier}  
\usepackage{url}  
\usepackage{graphicx}  
\usepackage{times}
\usepackage{amsmath}
\usepackage{latexsym}
\usepackage{amsfonts}
\usepackage{url}
\usepackage{algorithm}
\usepackage{algorithmic}
\usepackage{graphicx}
\usepackage{multirow}
\usepackage{enumitem}

\usepackage{color,url}
\usepackage{hyperref}

\newtheorem{Def}{Definition}
\begin{document}

\title{Goal-oriented Dialogue Policy Learning from Failures}
\author{Keting Lu$^1$, Shiqi Zhang$^2$, Xiaoping Chen$^1$\\
$^1$ School of Computer Science, University of Science and Technology of China\\
$^2$ Department of Computer Science, SUNY Binghamton\\
{\tt ktlu@mail.ustc.edu.cn; szhang@cs.binghamton.edu; xpchen@ustc.edu.cn}
}
\maketitle

\begin{abstract}

Reinforcement learning methods have been used for learning dialogue policies.
However, learning an effective dialogue policy frequently requires prohibitively many conversations.
This is partly because of the sparse rewards in dialogues, and the very few successful dialogues in early learning phase.
Hindsight experience replay (HER) enables learning from failures, but the vanilla HER is inapplicable to dialogue learning due to the implicit goals.
In this work, we develop two complex HER methods providing different trade-offs between complexity and performance, and, for the first time, enabled HER-based dialogue policy learning.
Experiments using a realistic user simulator show that our HER methods perform better than existing experience replay methods (as applied to deep Q-networks) in learning rate.


\end{abstract}


\section{Introduction}
\label{sec:intro}

Goal-oriented dialogue systems aim at assisting users to accomplish specific goals using natural language, and have been used in a variety of applications~\cite{zhang2015corpp,su2016line,li2017end}.  
Goal-oriented dialogue systems usually aim at \emph{concise} conversations.\footnote{In comparison, the Chatbots that want to maximize social engagement, such as Microsoft XiaoIce, frequently result in extended conversations.} 
Such dialogue systems typically include a language understanding component for recognizing and parsing the language inputs into inner representations, a belief state tracking component for predicting user intent and updating the dialogue history, and a dialogue management component that generates dialogue actions.
The dialogue actions can be converted into spoken or text-based language using a language generator.
Goal-oriented dialogue managers are frequently modeled as a sequential decision-making problem~\cite{young2013pomdp}, where reinforcement learning (RL) \cite{sutton1998reinforcement} can be used for learning an optimal dialogue policy from user experiences.
While a variety of RL methods have been developed for learning dialogue policies~\cite{williams2016end,cuayahuitl2017simpleds}, the methods typically require a large amount of dialogue experience until one can learn a good-quality dialogue policy.
In particular, successful dialogues are rare in early learning phase, making it a challenge for many dialogue systems to learn much at the beginning and producing poor user experiences.
This paper focuses on runtime dialogue data augmentation (DDA) for speeding up the process of learning goal-oriented dialogue policies.


\begin{figure*}[t]
\begin{center}
\vspace{-2em}
\includegraphics[width=0.7\textwidth]{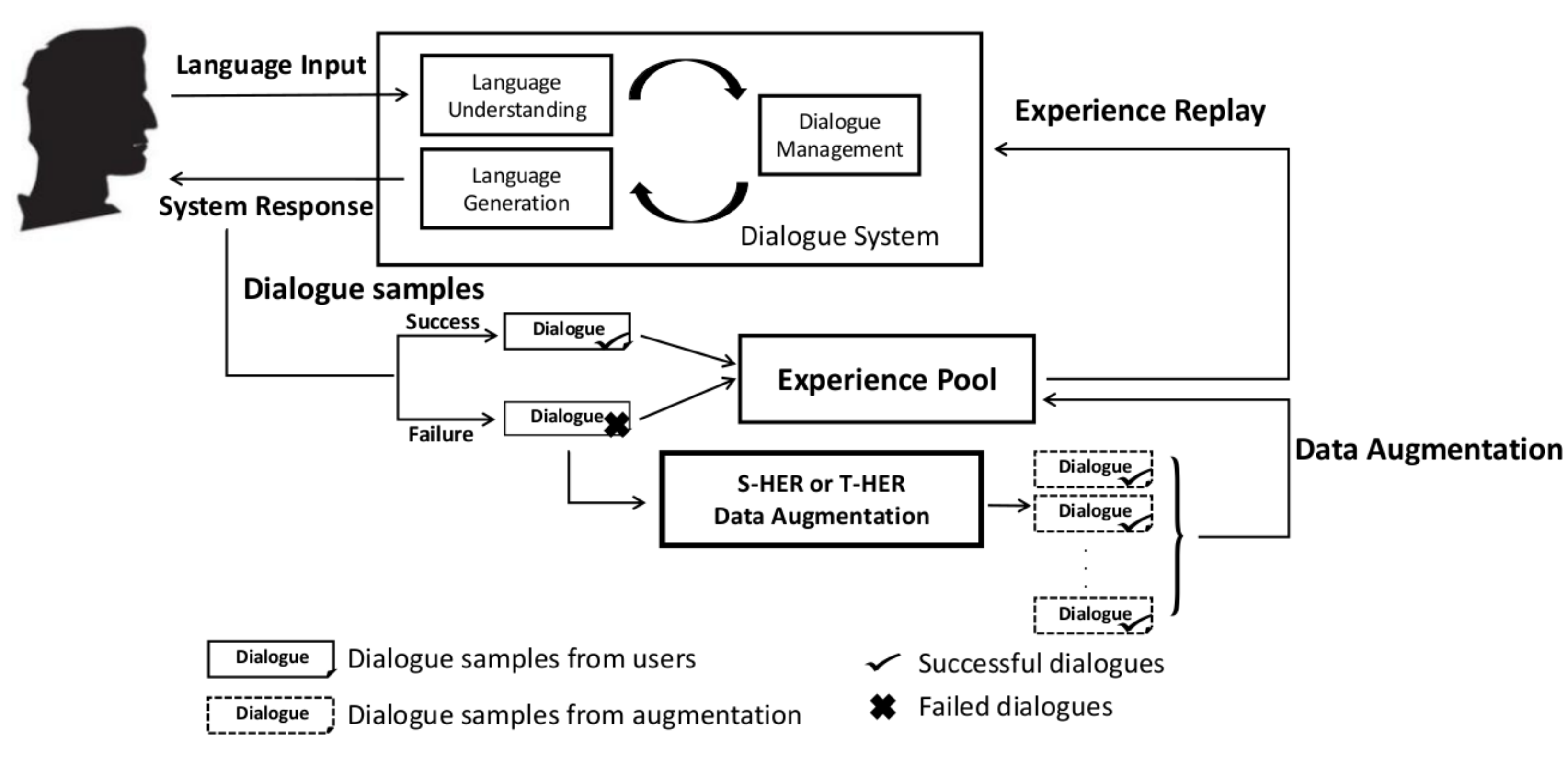}
\vspace{-1em}
\caption{Overview of our dialogue data augmentation (DDA) framework, and our two complex hindsight experience replay (HER) methods. }
\end{center}
\label{fig:complex_her}
\end{figure*}

The idea of data augmentation for RL tasks is not new. Existing research on hindsight experience replay (HER) has shown that, in robotic manipulation tasks, an RL agent's learning rate can be improved by changing goals to generate ``successful'' instances~\cite{andrychowicz2017hindsight}.
However, their HER approach is not directly applicable to dialogue domains, because a dialogue agent has to interact with people to identify the desired goal state (whereas the goal state is explicit in manipulation tasks).
We develop two complex HER methods for runtime DDA problems.
The first method is called Trimming-based HER (\textbf{T-HER}), where
failed dialogues are trimmed to generate successful dialogue instances.
While T-HER enables the agent to significantly augment the training data, the generated instances are relatively short.
The other method is Stitching-based HER (\textbf{S-HER}) that enables an agent to analyze the similarity between dialogue belief states, and stitch together dialogue segments to form relatively long, successful dialogues.

Figure~\ref{fig:complex_her} is an overview of our DDA framework, illustrating the role of our HER methods.
Like standard experience replay, we add user dialogues to the experience pool.
Unlike existing methods, when incoming dialogues are unsuccessful, we use our HER methods to generate new, successful dialogues, enabling our agent to learn from failures.

We have applied our DDA framework, including T-HER and S-HER, to Deep Q-Networks (DQNs)~\cite{mnih2015human} for learning dialogue policies. Results collected using a realistic user simulator~\cite{li2016user} suggest that our methods perform better than competitive experience replay baselines, including prioritized experience replay~\cite{schaul2015prioritized}.
Finally, our two HER methods can be combined to further improve the performance.
To the best of our knowledge, this is the first work on HER-based dialogue policy learning.


\section{Related Work}
\label{sec:related}

This work is closely related to research areas that aim at efficient (deep) RL methods, including, experience replay, reward shaping, exploration in RL, and supervised pre-training for RL.

Deep Q-Network (DQN) has enabled an agent to achieve human-level performance on playing Atari games~\cite{mnih2015human}.
RL algorithms, such as DQN, are generally data intensive and frequently require huge numbers of interactions with the environments. To better use the interaction experience, experience replay~(ER), that suggests storing and reusing samples at training time, has been widely used for speeding up the RL agent's training process~\cite{nair2015massively,wang2016sample}.
Prioritized ER (PER) further accelerates training by assigning a weight based on temporal difference error (TD-error) to each sample~\cite{schaul2015prioritized}, so as to increase the likelihood of selecting samples with a high TD-error.
While PER enables more effective sample selections~\cite{han2017multi}, the applicability of PER is limited when very few successful samples are available.
We develop HER methods that generate artificial ``successful'' samples to improve the learning rate of dialogue agents.

Reward shaping has been used for speeding up the learning of dialogue policies by adding a new dense reward function~\cite{ferreira2015reinforcement}. It has been proved that a well designed (dense) reward function does not alter the optimal policy~\cite{ng1999policy}. Recently, \citeauthor{su2015reward} applied Recurrent Neural Networks (RNNs) to predict the intermediate rewards for dialogues to reduce training time, and developed an on-line learning framework where dialogue policy and reward function were jointly trained via active learning~\cite{su2016line}. However, generating such complex reward functions require either considerable human effort or large datasets.



Another line of research focuses on developing effective exploration strategies for RL algorithms, enabling more sample-efficient dialogue learning.
For instance, \citeauthor{pietquin2011sampleA} integrated Least-Squares Policy Iteration~\cite{lagoudakis2003least} and
Fitted-Q~\cite{chandramohan2010optimizing}
for dialogue policy optimization~\cite{pietquin2011sampleA}.
Other examples include Gaussian Process (GP)-based sample-efficient dialogue learning~\cite{gasic2014gaussian},  
the Bayes-by-Backprop Q-network (BBQN)~\cite{lipton2017bbq},
and trust-region and gradient-based algorithms~\cite{su2017sample}.
Our HER methods have the potential to be combined with the above sample-efficient RL methods to produce a more efficient learning process.

Pre-training has been used in dialogue learning for computing an initial policy from a corpus using supervised learning (SL)~\cite{su2016continuously,peng2017composite}.         
After that, a dialogue agent can further improve the policy via learning from interactions with users.
In line with past research on dialogue systems (as listed above), we use pre-training in this work (unless stated otherwise) to give our dialogue agent a ``warm start''.

Work closest to this research is the original HER method~\cite{andrychowicz2017hindsight} that manipulates goals based on resulting states. But that method is only applicable to domains where goals are explicit to the agents, e.g., target positions in manipulation tasks. In dialogue domains, goals are not fully observable and must be identified via language actions, which motivates the development of complex HER methods in this work.


\section{Background}
In this section, we briefly introduce the two building blocks of this work, namely Markov decision process (MDP)-based dialogue management, and Deep Q-Network (DQN).


\subsection{MDP-based Dialogue Management}

Dialogue control is modeled using MDPs in this work.
An MDP-based dialogue manager~\cite{lipton2017bbq} can be described as a tuple $\langle \mathcal{S}, \mathcal{A}, T, s_0, \mathcal{R} \rangle$.
$\mathcal{S}$ is the state set, where $s\! \in\! \mathcal{S}$ represents the agent's current dialogue state including the agent's last action, the user's current action, the distribution of each slot, and other domain variables as needed.
$\mathcal{A}$ is the action set, where $a\! \in\! \mathcal{A}$ represents the agent's response.
$T$ is the stationary, probabilistic transition function with conditional density $p(s_{t+1}|s_t, a_t)$ that satisfies the (first-order) Markov property.
$s_0\! \in\! \mathcal{S}$ is the initial state.
$\mathcal{R}$: $\mathcal{S}\! \times\! \mathcal{A} \rightarrow \mathbb{R}$ is the reward function, where the agent receives a big bonus in successful dialogues, and has a small cost in each turn.

Solving an MDP-based dialogue management problem produces $\pi$, a dialogue policy. A dialogue policy maps a dialogue state to an action,
$\mathnormal{\pi}$: $\mathcal{S}\!\! \rightarrow\!\! \mathcal{A}$, toward maximizing the discounted, accumulative reward in dialogues, i.e., $R_t = \sum_{i=t}^{\infty}{\gamma^{i-t}r_{i}}$, where $\gamma\! \in\! [0,1]$ is a discount factor that specifies how much the agent favors future rewards.



\subsection{Deep Q-Network}
Deep Q-Network (DQN)~\cite{mnih2015human} is a model-free RL algorithm for discrete action space. DQN uses a neural network as an approximation of the optimal Q-function, $Q^\ast=Q(s,a;\theta)$, where $a$ is an action executed at state $s$, and $\theta$ is a set of parameters. Its policy is defined either in a $greedy$ way: $\pi_Q(s)=\arg \max_{a \in \mathcal{A}}Q(s,a;\theta)$ or being $\epsilon\!-\!greedy$, i.e., the agent takes a random action in probability $\epsilon$ and action $\pi_Q(s)$ otherwise. The loss function for minimization in DQN is usually defined using TD-error:
\begin{equation}\label{loss_function}
  \mathcal{L} = \mathbb{E}_{s,a,r,s^\prime}[(Q(s,a;\theta)-y)^2],
\end{equation}
where $y=r+\gamma \max_{a^\prime \in \mathcal{A}}Q(s^\prime,a^\prime;\theta)$.

Naive DQNs often suffer from overestimation and instability, and two techniques are widely used  to alleviate the issues.
One is called \emph{target network}~\cite{mnih2015human} whose parameters are updated by $\theta$ once every many iterations in the training phase.
The other technique is \emph{experience replay}~\cite{lin1993reinforcement,mnih2015human}, where an experience pool $\mathcal{E}$ stores samples, each in the form of $(s_t, a_t, r_t, s_{t+1})$.
In training the DQN, a mini-batch is uniformly sampled from $\mathcal{E}$. Experience replay desensitizes DQN to the correlation among the samples, and increases the data efficiency via reusing the (potentially expensive) samples. Both techniques improve the performance of DQN and are used in this paper.


\section{Dialogue Segmentation}
\label{sec:segment}

In this section, we introduce the concept of \emph{dialogue subgoal}, and present a segmentation algorithm that efficiently outputs \emph{valid} dialogue segments, which are later used in our complex HER methods.


\subsection{Definition of Dialogue Subgoal}
\label{sec:subgoal}

Goal-oriented dialogue systems help users accomplish their goals via dialogue, where a goal $G$ includes a set of constraints $C$ and a set of requests $R$~\cite{schatzmann2009hidden}: $G=(C,R)$.

Consider a movie booking domain.
A user may ask about the \emph{name} and \emph{time} of a movie starring \emph{Jackie Chan}, of genre \emph{action}, and running \emph{today}, where the goal is in the form of:
\begin{equation*}\label{eqn:goal}
\footnotesize
\begin{aligned}
\textbf{Goal}= \Bigg( & C=\begin{bmatrix}
 actor = Jackie~Chan\\
 genre = action \\
 date = today
 \end{bmatrix} \textbf{,}
\\
&R=\begin{bmatrix}
 movie~name = \\
 start~time =
 \end{bmatrix}
\Bigg )
\end{aligned}
\end{equation*}



\begin{Def}
\textbf{Subgoal}\footnote{Our definition of \emph{subgoal} is different from that in the dialogue learning literature on hierarchical reinforcement learning~\cite{tang2018subgoal} and state space factorization~\cite{thomson2013statistical}.} 
Given $G=(C,R)$ and $G'=(C',R')$, we say $G'$ is a subgoal of $G$, or $G'\sqsubset G$, if $C' \subset C$ and $R' \subset R$, where $G' \neq \emptyset $ .
  \label{def:subgoal}
\end{Def}
In successful dialogues, the agent must correctly identify all constraints and requests in $G$, so as to correctly provide the requested information via querying a database.
Figure~\ref{fig:subgoals} shows three example subgoals (out of many) in the movie booking example.

\begin{figure}[t]
\vspace{-.5em}
\begin{equation*}\label{eqn:subgoals}
\footnotesize
\begin{aligned}
\textbf{Subgoal 1}= \Bigg( & C=\begin{bmatrix}
 actor = Jackie~Chan\\
 genre = action
 \end{bmatrix} \textbf{,}~~R=\emptyset\Bigg ).
\\
\textbf{Subgoal 2}= \Bigg( & C=\emptyset \textbf{,}~~R=\begin{bmatrix}
 movie~name = \\
 start~time =
 \end{bmatrix}
\Bigg ).
\\
\textbf{Subgoal 3}= \Bigg( & C=\begin{bmatrix}
 actor = Jackie~Chan\\
 date = today
 \end{bmatrix} \textbf{,}\\
 &R=\begin{bmatrix}
 movie~name = \\
 start~time =
 \end{bmatrix}
\Bigg ).
\end{aligned}
\end{equation*}
\vspace{-1em}
\caption{Example subgoals in the movie booking domain. For instance, {\bf Subgoal 3} corresponds to the request of ``\emph{Please tell me the name and start time of the movie that is playing today and stars Jackie Chan. }''}
\label{fig:subgoals}
\end{figure}


It should be noted that some subgoals do not make sense to humans, but can be useful for dialogue learning.
For instance, Subgoal 2 corresponds to a query about \emph{movie name} and \emph{start time} without any constraints. Real users do not have such goals, but an agent can still learn from the experience of achieving such subgoals.

Continuing the ``Jackie Chan'' example, if the agent misidentifies \emph{genre}, \emph{start time}, or both, the dialogues will be deemed unsuccessful, meaning that the agent cannot learn much from it, even though the agent has correctly identified the other entries of \emph{movie name}, \emph{actor}, etc. In this work, we make use of such unsuccessful dialogues in the training process, leveraging the fact that the agent has achieved subgoals (not all) in these dialogues.



\subsection{Dialogue Segment Validation}

Given dialogues $\mathcal{D}'$ and $\mathcal{D}$, we say $\mathcal{D}'$ is a segment of $\mathcal{D}$, if $\mathcal{D}'$ includes a consecutive sequence of turns of $\mathcal{D}$.
We introduce an assessment function\footnote{Such assessment functions are provided by dialogue simulators, e.g., TC-Bot (https://github.com/MiuLab/TC-Bot).}, $success(G, \mathcal{D})$, that outputs \emph{true} or \emph{false} representing whether dialogue $\mathcal{D}$ accomplishes goal (or subgoal) $G$ or not.
Using the assessment function, we define the validity of dialogue segments.

 \begin{Def}
 \textbf{Validity} of dialogue segments:
   Given dialogue $\mathcal{D}$, goal $G$, and dialogue segment $\mathcal{D}'$ (of $\mathcal{D}$), we say $\mathcal{D}'$ is a valid dialogue segment of $\mathcal{D}$, iff there exists a subgoal $G' \sqsubset G$, and $\mathnormal{success}(G',\mathcal{D}')$ is \emph{true}.
   \label{dialoguesegment}
 \end{Def}

Using Definitions~\ref{def:subgoal} and~\ref{dialoguesegment}, one can assess the validity of dialogue segment $D'$, using the entire dialogue $D$, the goal of this dialogue $G$, and the provided subgoal $G'$.
However, there exist many subgoals of the ultimate goal (combinatorial explosion), and it soon becomes infeasible to assess the validity of a dialogue segment.
Formally, given goal $G=(C,R)$, the number of subgoals is
$\sum_{i=0}^{\vert C \vert}{\binom{i}{\vert C \vert}}\cdot \sum_{j=0}^{\vert R \vert}{\binom{j}{\vert R \vert}} - 2$, where we subtract the two extreme cases of the subgoal being $\emptyset$ and $G$.
If $\vert C \vert = 5$ and $\vert R \vert = 5$, the number of subgoals is 1598.

\subsection{Dialogue Segmentation Algorithm}

Instead of assessing the validity of a dialogue segment using \emph{all} subgoals, we aim at using only the ones with the so-far-highest ``cardinality''.
The intuition comes from the following two observations.
In line with previous research on dialogue policy learning, e.g.,~\cite{schatzmann2007agenda}
, we assume users are cooperative and consistent in dialogues.

\begin{enumerate}
\item During a dialogue, the number of constraints and requests that have been identified is monotonically increasing; and

\item When a subgoal is accomplished, all its ``subsubgoals'' are automatically accomplished.
\end{enumerate}

\begin{algorithm}[t]
\caption{Dialogue Segmentation}
\label{alg1}
\begin{algorithmic}[1] \footnotesize
\REQUIRE ~~\\
Goal $G$ that includes constraint set $C$ and request set $R$; \\
Assessment function $\mathnormal{success(\cdot,\cdot)}$;
\ENSURE ~~\\
A collection of pairs of a valid (head) dialogue segment and a corresponding subgoal, $\Omega$;

\STATE Initialize $\Omega=\emptyset$
\STATE Initialize $\mathbb{P}=\emptyset$ and $\mathbb{Q}=C \cup R$
\WHILE{Dialogue $\mathcal{D}$ is not ended}
\STATE Outcome flag $segment\_outcome = False$
\FOR{$q \in \mathbb{Q}$}
\STATE Construct a subgoal $G' = \mathbb{P} \cup q$
\IF{$\mathnormal{success}(G',\mathcal{D})$}
\STATE $segment\_outcome = True$
\STATE $\mathbb{P} \gets \mathbb{P} \cup q$ and $\mathbb{Q} \gets \mathbb{Q}\setminus q$
\ENDIF
\ENDFOR
\IF{$segment\_outcome == True$}
\STATE $ \Omega \gets \Omega \, \cup <\!\!D,\mathbb{P}\!\!> $
\ENDIF
\ENDWHILE
\end{algorithmic}
\end{algorithm}

Leveraging the above two observations, we develop a \emph{dialogue segmentation} algorithm to efficiently identify valid dialogue segments.
First, we initialize two collections $\mathbb{P}$ and $\mathbb{Q}$.
$\mathbb{P}$ stores the constraints and requests that have been identified during the dialogues, and $\mathbb{Q}$ stores the ones that have not been identified.
Therefore, at the very beginning of dialogues, $\mathbb{P}$ is an empty set, and $\mathbb{Q}$ stores all constraints and requests.
After each dialogue turn, we generate a subgoal set $\mathbb{G}$ where all subgoals share the same (so far highest) cardinality. Formally, $\mathbb{G}=\{G'|G'=\mathbb{P} \cup q, \forall q \in \mathbb{Q}\}$. If $G' \in \mathbb{G}$ is accomplished by the current dialogue segment, the corresponding $q$ is removed from $\mathbb{Q}$ and added into $\mathbb{P}$, which is used to generate the subgoal set for the next dialogue segment. So at each dialogue turn, only a small set of subgoals (i.e., $\mathbb{G}$) is used to assess a dialogue segment. More detailed procedure is shown in Algorithm~\ref{alg1}. Compared to exhaustive subgoal identification that suffers from combinatorial explosion, our dialogue segmentation algorithm has $O(|C|+|R|)$ time complexity.

Building on our dialogue segmentation algorithm, we next introduce our runtime data augmentation methods for efficient dialogue policy learning.


\section{Our Complex HER Methods for Dialogue Data Augmentation}
\label{sec:dda}


In this section, we introduce two complex hindsight experience replay (HER) methods for dialogue data augmentation (DDA).
The original HER method~\cite{andrychowicz2017hindsight} is not applicable to dialogue domains, because goals in dialogues are not explicitly given (c.f., path planning for robot arms) and must be identified in dialogue turns.
We say our HER methods are ``complex'', because our methods manipulate dialogue experiences based on the resulting dialogue states (which is more difficult than goal manipulation).
The HER methods generate successful, artificial dialogues using dialogues from users (successful or not), and are particularly useful in early learning phase, where successful dialogues are rare.


\subsection{Trimming-based HER (T-HER)}
\label{sec:trim}

The idea of T-HER is simple: we pair valid dialogue segments (from Algorithm~\ref{alg1}) and their corresponding subgoals to generate successful dialogue instances for training.

It requires care in rewarding the success of the generated dialogues (c.f., the ones from users).
We want to encourage the agent to accomplish subgoals (particularly the challenging ones) using positive rewards, while avoiding the agent sticking to accomplishing only the subgoals.
Given $R_{max}$ and $R_{min}$ being the reward and penalty to successful and unsuccessful dialogues with users, we design the following (simple) reward function for the successful, artificial dialogues from T-HER.
%
\begin{equation}\label{eqn:segment}
R(\mathcal{D})= \alpha \cdot |G'|,~\textnormal{ensuring}~~\alpha \cdot |G'| < R_{max}
\end{equation}
where $\alpha$ is a weight, $G'$ is a subgoal, $\mathcal{D}$ is a dialogue segment, and $|G'|$ is the number of entries of $G'$ that must be identified. The agent receives a small penalty (-1 in our case) at each turn to encourage short dialogues.



T-HER enables a dialogue agent to learn from the experience of completing imaginary, relatively easy tasks, and is useful while accomplishing ultimate goals is challenging (e.g., in early learning phase).
However, as a relatively simple strategy for DDA, T-HER cannot generate dialogues that are more complex that those from users.


\subsection{Stitching-based HER (S-HER)}
\label{sec:stitch}

S-HER is relatively more complex than T-HER, while both need valid dialogue segments from Algorithm~\ref{alg1}.
T-HER uses the segments as successful dialogues that accomplish subgoals, whereas S-HER uses them as parts to construct new, successful dialogues.
The key questions to S-HER are \emph{what dialogue segments are suitable for stitching,} and \emph{how they are stitched together}.
Next, we define \emph{stitchability} of dialogue segments using \emph{KL Divergence}:
$$
D_{K\!L}(\mathit{s_0} \vert \vert \mathit{s_1})= \sum_{i=0}^{n}{\mathit{s_0}(\mathnormal{i}) \cdot \log \frac{\mathit{s_0}(\mathnormal{i})}{\mathit{s_1}(\mathnormal{i})}}
$$
where $s_0$ and $s_1$ are two probability distributions.

\begin{Def}
 \label{def:stitch}
 \textbf{Stitchability:}
  Consider two dialogue segments
  $$
  	\mathcal{D} = { (s_0,a_0,s_1), (s_1,a_1,s_2), \cdots , (s_{M-1},a_{M-1},s_{M})},
  $$
  $$
  	\mathcal{D}' = {(s'_0,a'_0,s'_1), (s'_1, a'_1, s'_2), \cdots , (s'_{N-1},a'_{N-1},s'_{N})},
  $$
where $M$ and $N$ are turn numbers (dialogue segment lengths) of $\mathcal{D}$ and $\mathcal{D}'$, and each turn includes initial state, dialogue action, and resulting state.

If the dialogues' corresponding (sub)goals are $G$ and $G'$, we say $\mathcal{D}$ and $\mathcal{D}'$ are stitchable, if and only if $G = G'$, and
  \label{stitch}
\end{Def}
  $$
  	D_{K\!L}(s_M ~\vert \vert~ s'_0) \leq \varepsilon,
  $$
  \emph{where
  $\varepsilon\! \in\! \mathbb{R}$ is a stitchability threshold.}
  \vspace{.5em}


\begin{algorithm}[t]
\caption{Stitching-based HER}
\label{stitchalg}  \footnotesize
\begin{algorithmic}[1]
\REQUIRE ~~\\
Dialogue $\mathcal{D}$, and its goal $G$ \\
A collection of pairs of a valid (tail) dialogue segment and a corresponding subgoal, $\Gamma$ \\
KL-divergence threshold, $\varepsilon$ \\
Assessment function, $success(\cdot,\cdot)$
\ENSURE ~~\\
A set of newly generated dialogues, $\mathbb{M}$
\STATE Initialize $\mathbb{M} \gets \emptyset$
\STATE Assess input dialogue: \emph{success\_flag} $= success(G,\mathcal{D})$
\STATE Call Algorithm~\ref{alg1} to compute $\Omega$, using $\mathcal{D}$, $G$, and $success(\cdot,\cdot)$, where $\Omega$ is a collection of pairs of a valid (head) dialogue segment and a corresponding subgoal
\FOR{$<\!\!\mathcal{D'},G'\! > \in \Omega$}
\IF{~\emph{success\_flag}~ \textbf{is} ~\emph{True}~}
\STATE $ \Gamma \gets \Gamma \, \cup <\!\!\mathcal{D} \ominus \mathcal{D'},G'\!\!>$, where $\ominus$ is a dialogue subtraction operator
\ELSE
\FOR{$ <\!\!\mathcal{D''},G''\!\!> \in \Gamma $}
\IF{$ G'~\textbf{is}~ G'' \, and \, D_{K\!L}(\mathcal{D'} \vert \vert \mathcal{D''}) \leq \varepsilon $}
\STATE $\mathcal{D}^{stitched} = concatenate(\mathcal{D}',\mathcal{D}'')$
\STATE $\mathbb{M} \gets \mathbb{M} \cup \mathcal{D}^{stitched}$
\STATE Break
\ENDIF
\ENDFOR
\ENDIF
\ENDFOR
\STATE Return $\mathbb{M}$
\end{algorithmic}
\end{algorithm}

We use \emph{head (tail) dialogue segment} to refer to the segment that includes the early (late) turns in the resulting dialogue. In Definition~\ref{def:stitch}, $\mathcal{D}$ and $\mathcal{D}$ are head and tail segments respectively.

The key idea of S-HER is to use unsuccessful user dialogues to generate valid \emph{head} dialogue segments (stored in set $\Omega$), use successful user dialogues to generate valid \emph{tail} dialogue segments (stored in set $\Gamma$), and stitch dialogue segments from the two sets (one from each set) to form new successful artificial dialogues, only if their connecting dialogue states are similar enough.


Algorithm~\ref{stitchalg} presents S-HER.
Given a new user dialogue $\mathcal{D}$, S-HER first calls Algorithm~\ref{alg1} to generate all valid dialogue segments that start from the beginning of $\mathcal{D}$, and stores them in set $\Omega$.
After that, S-HER assesses $\mathcal{D}$'s successfulness using its associated goal, and stores the result in \emph{success\_flag}.
If successful, S-HER subtracts segments in $\Omega$ from $\mathcal{D}$ to generate the tail segments that eventually lead to successful dialogues. The tail segments are used to augment $\Gamma$.
$\ominus$ in Line 6 is a dialogue subtraction operator, i.e., $\mathcal{D} \ominus \mathcal{D}'$ produces a dialogue segment by removing $\mathcal{D}'$ from $\mathcal{D}$, and this operation is valid, only if $\mathcal{D}'$ is a segment of $\mathcal{D}$.
If the new user dialogue is unsuccessful, S-HER concatenates one head segment from $\Omega$ to one tail segment from $\Gamma$ to form a new, successful dialogue $\mathcal{D}^{stitched}$, and add it into set $\mathbb{M}$, which is used for storing the algorithm output.

\vspace{1em}
\noindent
\textbf{Remarks: } S-HER uses KL divergence (Line 9 in Algorithm~\ref{stitchalg}) to measure the similarity between dialogue states. If their divergence is lower than a threshold, S-HER stitches the corresponding two dialogue segments together to generates a new dialogue.
S-HER is more effective than T-HER, when there are very few successful dialogues in the experience replay pool.
This is because S-HER is able to generate many successful dialogues, even if there is only one successful dialogue from user, which significantly augments the dialogue data for accelerating the training process.

\section{Experiment}


Experiments have been conducted to evaluate the key hypothesis of T-HER and S-HER being able to improve the learning rate of DQN-based dialogue policies. We have combined the two methods with prioritized experience replay~\cite{schaul2015prioritized} to produce better results, and compared T-HER and S-HER under different conditions.



\subsection{Dialogue Environment}
\label{SecExp1}
Our complex HER methods were evaluated using a dialogue simulation environment, where a dialogue agent communicates with simulated users on \emph{movie-booking} tasks~\cite{li2016user,li2017end}.
This movie-booking domain consists of 29 slots of two types, where one type is on \emph{search constraints} (e.g., number of people, and date), and the other is on \emph{system-informable} properties that are needed for database queries (e.g., critic rating, and start time).

A dialogue state consists of five components:
1) one-hot representations of the current user action $act$ and mentioned slots;
2) one-hot representations of last system action $act$ and mentioned slots;
3) the belief distribution of possible values for each slot;
4) both a scalar and one-hot representation of the current turn number; and
5) a scalar representation indicating the number of results which can be found in a database according to current search constraints.

The system (dialogue agent) has 11 dialogue actions representing the system intent. These actions can be mapped into a natural language response according to the dialogue belief states and heuristic rules. Once a dialogue reaches the end, the system receives a big bonus $80$, if it successfully helps users book tickets. Otherwise, it receives $-40$.
In each turn, the system receives a fixed reward $-1$ to encourage shorter dialogues. The maximum number of turns allowed is 40.
Our previous work has studied how such rewards affect dialogue behaviors, e.g., higher success bonus and/or lower failure penalty encourage more risk-seeking behaviors and higher QA costs result in less accurate, shorter conversations~\cite{zhang2015corpp}.

The DQN architecture used in this paper is a single-layer neural network of size 80. Its output layer has 11 units corresponding to the 11 dialogue actions.
The techniques of target network and experience replay are applied (see Section~\ref{sec:related}).
The size of experience pool is $100k$, and experience replay strategy is uniform sampling.
The value of $\alpha$ in Equation~\ref{eqn:segment} is 1.0, and
$\epsilon\!-\!greedy$ policy is used, where $\epsilon$ is initialized with 0.3, and decayed to 0.01 during training.

Each experiment includes 1000 epochs. Each epoch includes 100 dialogue episodes. By the end of each epoch, we update the weights of target network using the current behavior network, and this update operation executes once every epoch.
Every data point in the figures is an average of 500 dialogues from 5 runs (100 dialogues in each run). For instance, a data point of 0.6 success rate at episode 70 means that 60\% of the 70th dialogue episodes (500 in total) were successful.
In line with previous research~\cite{su2016continuously,peng2017composite}, we use
supervised learning to give our dialogue agent a ``warm start'' in each set of experiments, unless specified otherwise.



\begin{figure}[t]
\vspace{-.5em}
\centering
\includegraphics[width=\columnwidth]{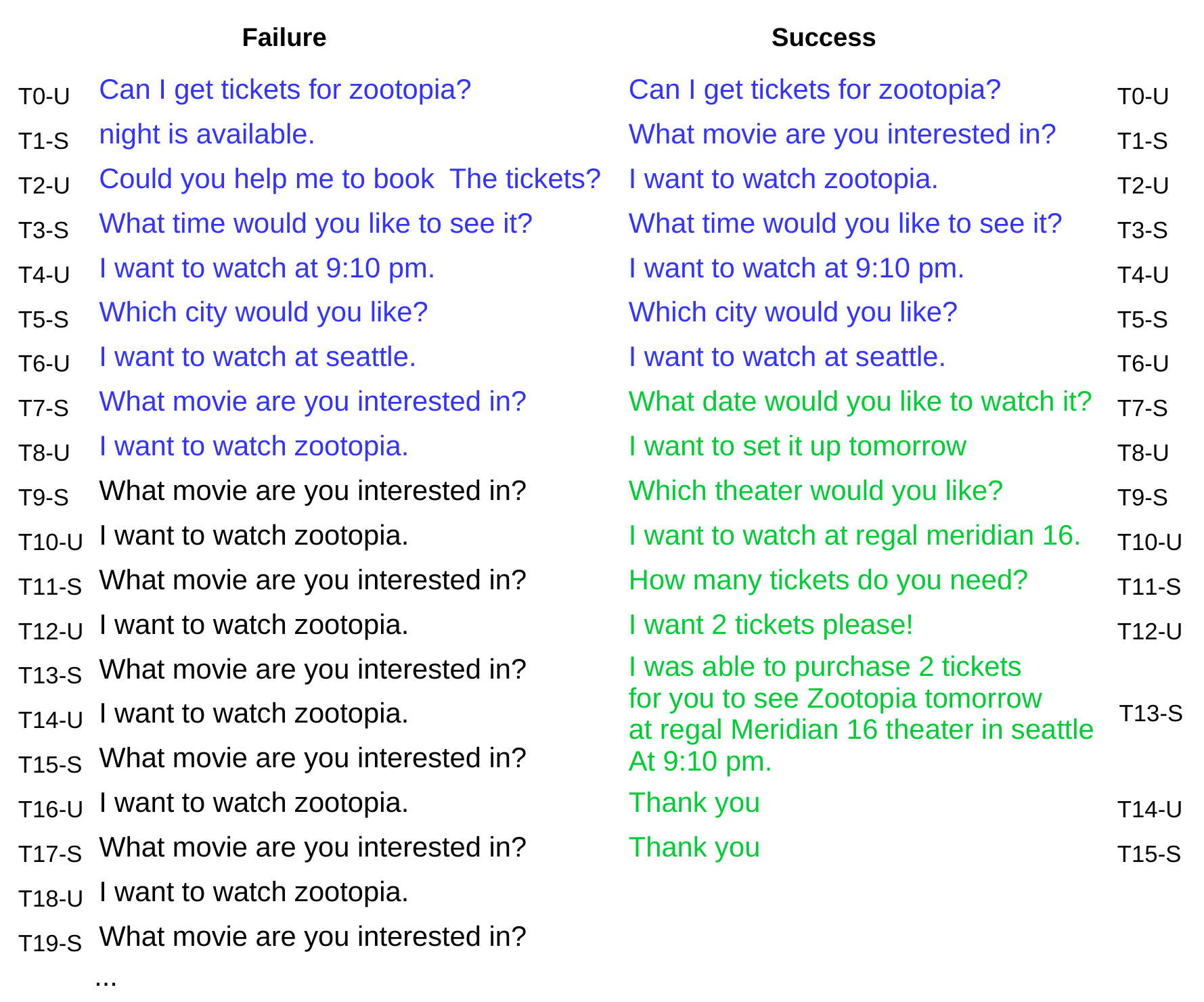}
\vspace{-2em}
\caption{Examples of an unsuccessful dialogue and a successful dialogue, where we use ``U'' and ``S'' to indicate \emph{user} and \emph{system} turns respectively. }
\label{fig:dialogueExample}
\end{figure}


\subsection{Case Illustration}
Before presenting statistical results, we use two example dialogues to illustrate how T-HER and S-HER augment dialogue data, as shown in Figure~\ref{fig:dialogueExample}.
The goal of the unsuccessful dialogue
(Left) includes three constraints of \emph{start time}, \emph{city}, and \emph{movie name}, and one request of \emph{ticket}.
The goal of successful dialogue (Right) includes six constraints of \emph{start time}, \emph{city}, \emph{movie name}, \emph{theater}, \emph{date}, and \emph{number of people}, and one request of \emph{ticket}. Next, we demonstrate how T-HER and S-HER use the two dialogues to generate artificial, successful dialogues.

T-HER selects valid dialogue segments
according to achieved subgoals.
On the left, the dialogue segment in blue color achieves the subgoal that includes three constraints (\emph{start time}, \emph{city}, and \emph{movie name}), and also its ``subsubgoals''.
Accordingly, T-HER can generate three valid dialogue segments (new, successful dialogues), including the ones from $T_0$ to $T_4$, from $T_0$ to $T_5$, and from $T_0$ to $T_8$.
Turns after $T_8$ are not considered, because our assessment function does not allow the agent asking useless questions.

S-HER generates new successful dialogues by stitching head and tail dialogue segments.
In this example, the two dialogue segments in blue color achieved the same subgoal, although the two dialogue segments have very different flows.
S-HER enables our agent to generate a new successful dialogue by stitching the dialogue segment in blue color on the left, and the dialogue segment in green color on the right.
The newly generated, successful dialogue achieves the goal that is originally from the dialogue on the right.

\begin{figure}[t]

\centering
\includegraphics[width=0.85\columnwidth]{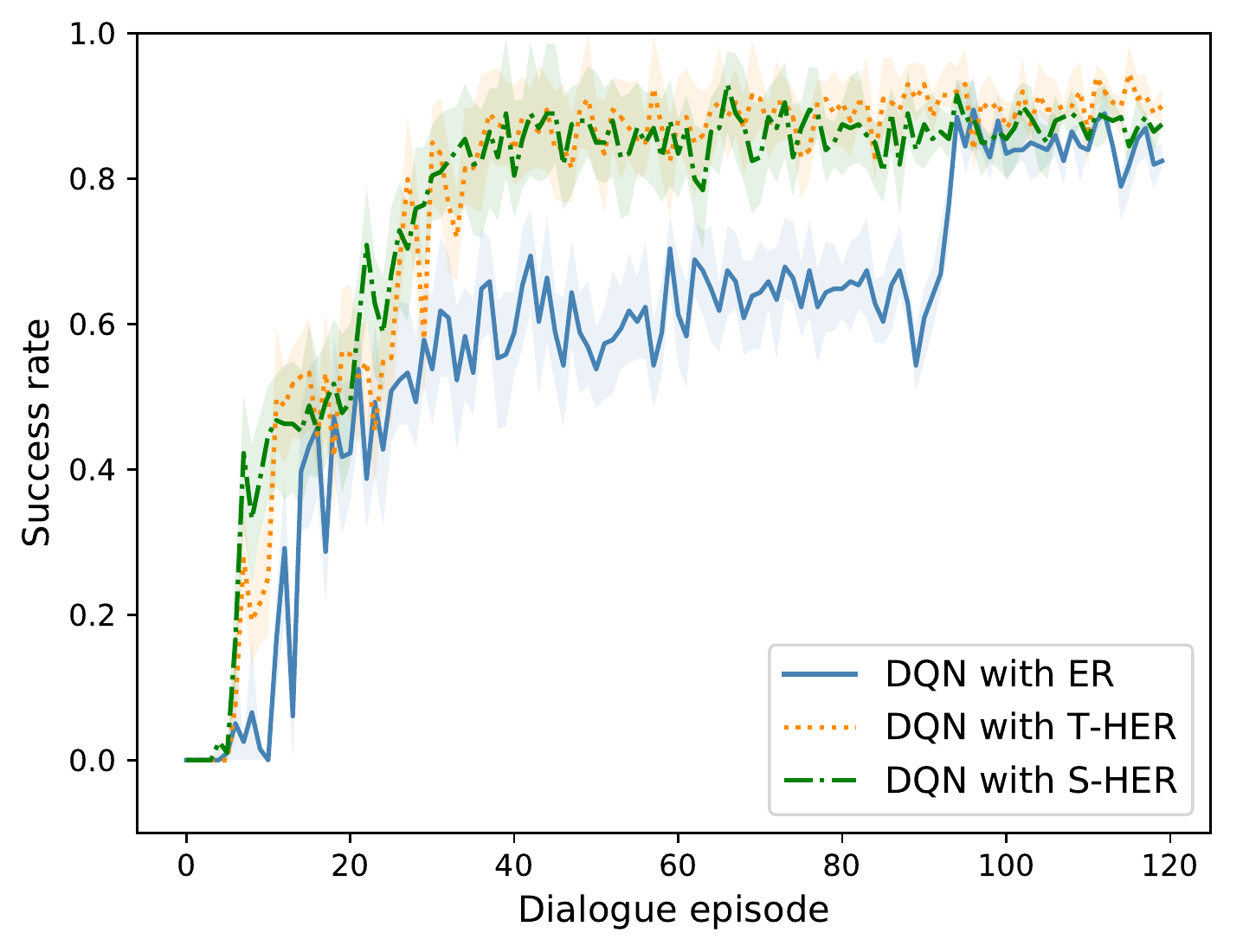}
\vspace{-1em}
\caption{Learning to dialogue to accomplish tasks of booking movie tickets. Our two complex HER strategies improve the learning rate, c.f., naive DQN with standard experience reply (ER). }
\label{fig:baseline}
\end{figure}

\subsection{Experimental Results}

Experiments have been extensively conducted to evaluate the following hypotheses.
I)~Our HER methods perform better than standard experience reply; II)~Our HER methods can be combined with prioritized ER for better performance; III)~Our HER methods perform better than existing ER methods in cold start; and IV)~T-HER and S-HER can be combined to produce the best performance.

\vspace{-1em}
\paragraph{DQN with our HER strategies}
\label{SecExp2}
Since the original HER method~\cite{andrychowicz2017hindsight} is not applicable to dialogue systems, we compare our complex HER methods with a standard DQN-based reinforcement learner.
Figure~\ref{fig:baseline} shows that both T-HER and S-HER significantly accelerate the training process, which supports our key hypothesis.
The KL divergence threshold of S-HER is 0.2 in this experiment. Next, we study the effect this parameter to the performance of S-HER.

\begin{figure}[t]

\centering
\includegraphics[width=0.85\columnwidth]{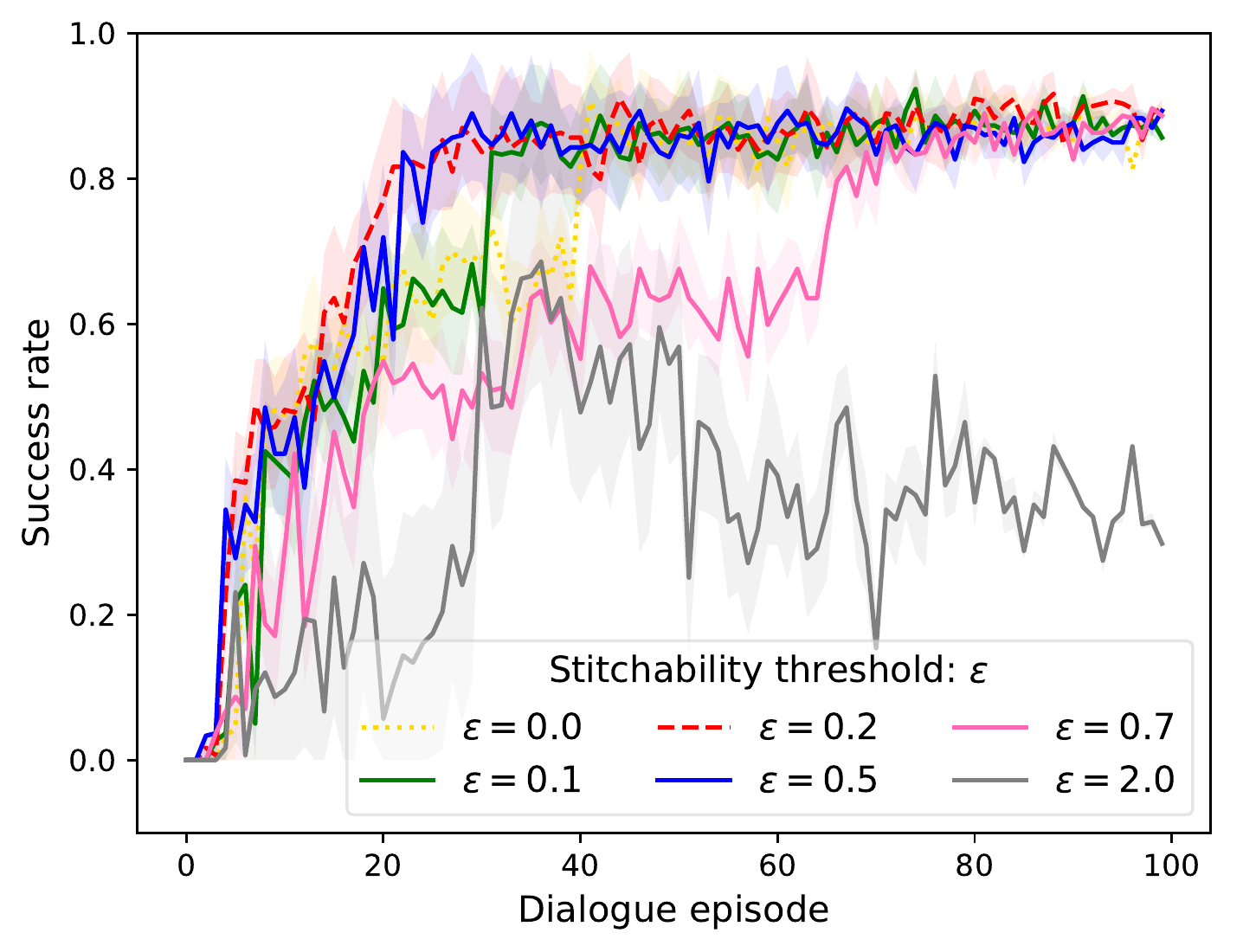}
\vspace{-1em}
\caption{\emph{Stitchability} in S-HER: KL divergence threshold $\varepsilon$ in [0.2, 0.5] performs the best.
}
\label{fig:thresholds}
\end{figure}

\vspace{-1em}
\paragraph{KL divergence level of S-HER}\label{SecExp3}
Stitching is allowed in S-HER, only if the KL divergence between two connecting states is below a ``stitchability'' threshold (details in Algorithm~\ref{stitchalg}).
Intuitively, if the threshold is too small, there will be very few dialogue segments being stitched (though the newly generated dialogues are of very good quality).
If the threshold is too big, many dialogue segments are stitched, but the quality of generated, artificial dialogues will be poor, producing negative effects to the learning process.

Figure~\ref{fig:thresholds} depicts the influences of different thresholds.
As expected, when the threshold is too high (e.g., $\varepsilon=2.0$), the RL agent failed to converge to a good policy; when the threshold is too small (e.g., $<=0.1$), the improvement to the learning rate is not as good as using a threshold within the range of $[0.2, 0.5]$.
Results here can serve as a reference to S-HER practitioners.

\begin{figure}[t]
\centering
\includegraphics[width=0.85\columnwidth]{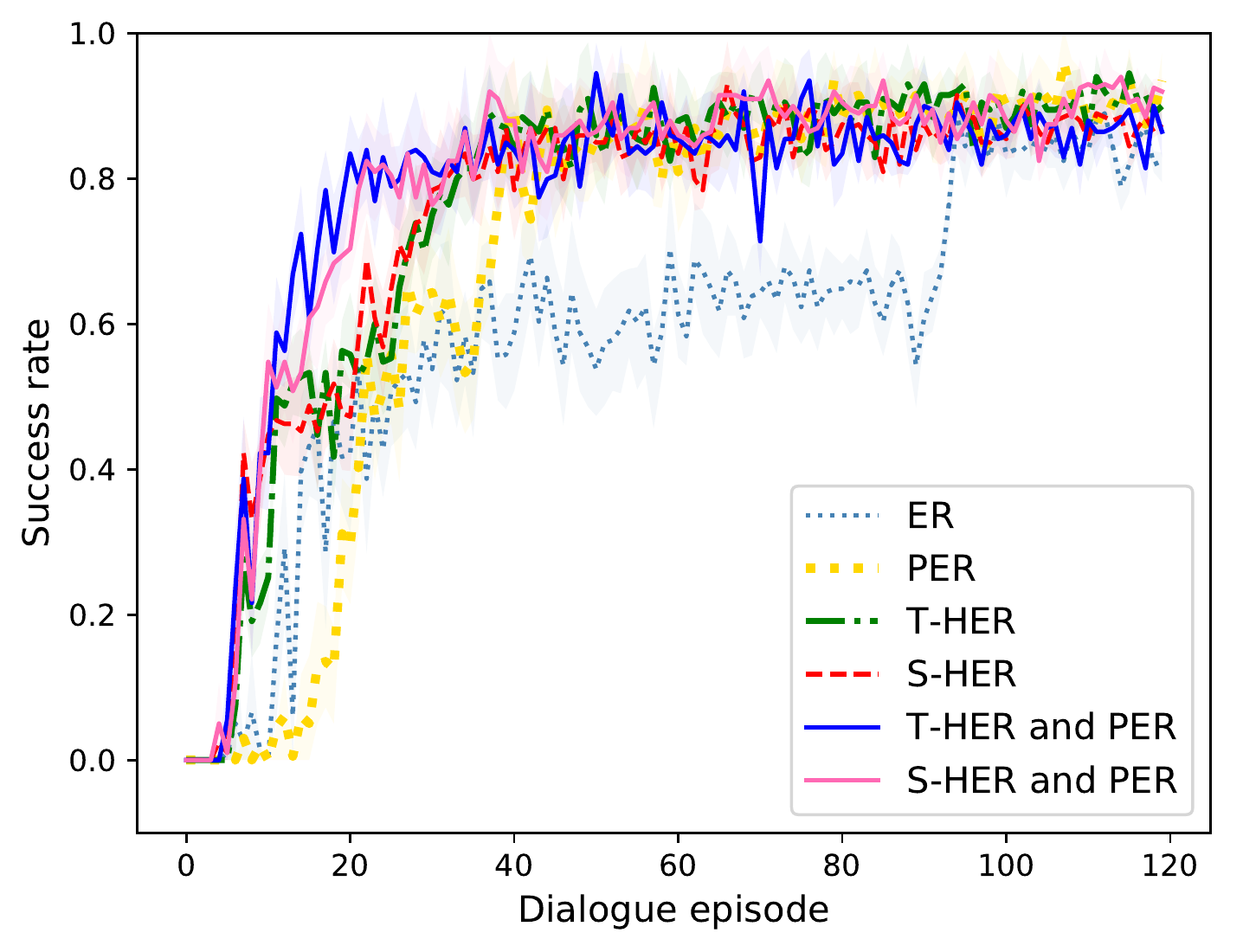}
\vspace{-1em}
\caption{Incorporating prioritized experience replay to improve the performances of T-HER and S-HER. }
\label{fig:combine}
\end{figure}

\vspace{-1em}
\paragraph{Integration with PER}\label{SecExp4}
The first two sets of experiments used uniform sampling in experience replay.
We further evaluate the performance of integrating our complex HER methods with prioritized experience replay (PER) that prioritizes selecting the potentially-valuable samples (in terms of TD-error)~\cite{schaul2015prioritized}.
Figure~\ref{fig:combine} shows the results. We can see T-HER and S-HER perform better than PER-only, and that incorporating PER into T-HER and S-HER further improved the performances.
It should be noted that all five learning strategies converge to policies that are of similar qualities.


\vspace{-1em}
\paragraph{Learning with a cold start}\label{SecExp5}

T-HER and S-HER produced similar results in previous experiments.
In this set of experiments, we removed the warm-start policy (from supervised learning) that has been widely used in the literature~\cite{li2017end,lipton2017bbq}, and reduced the experience replay pool from $100k$ to $1k$, resulting in an extremely challenging task to the dialogue learners.

\begin{figure}[t]

\centering
\includegraphics[width=0.85\columnwidth]{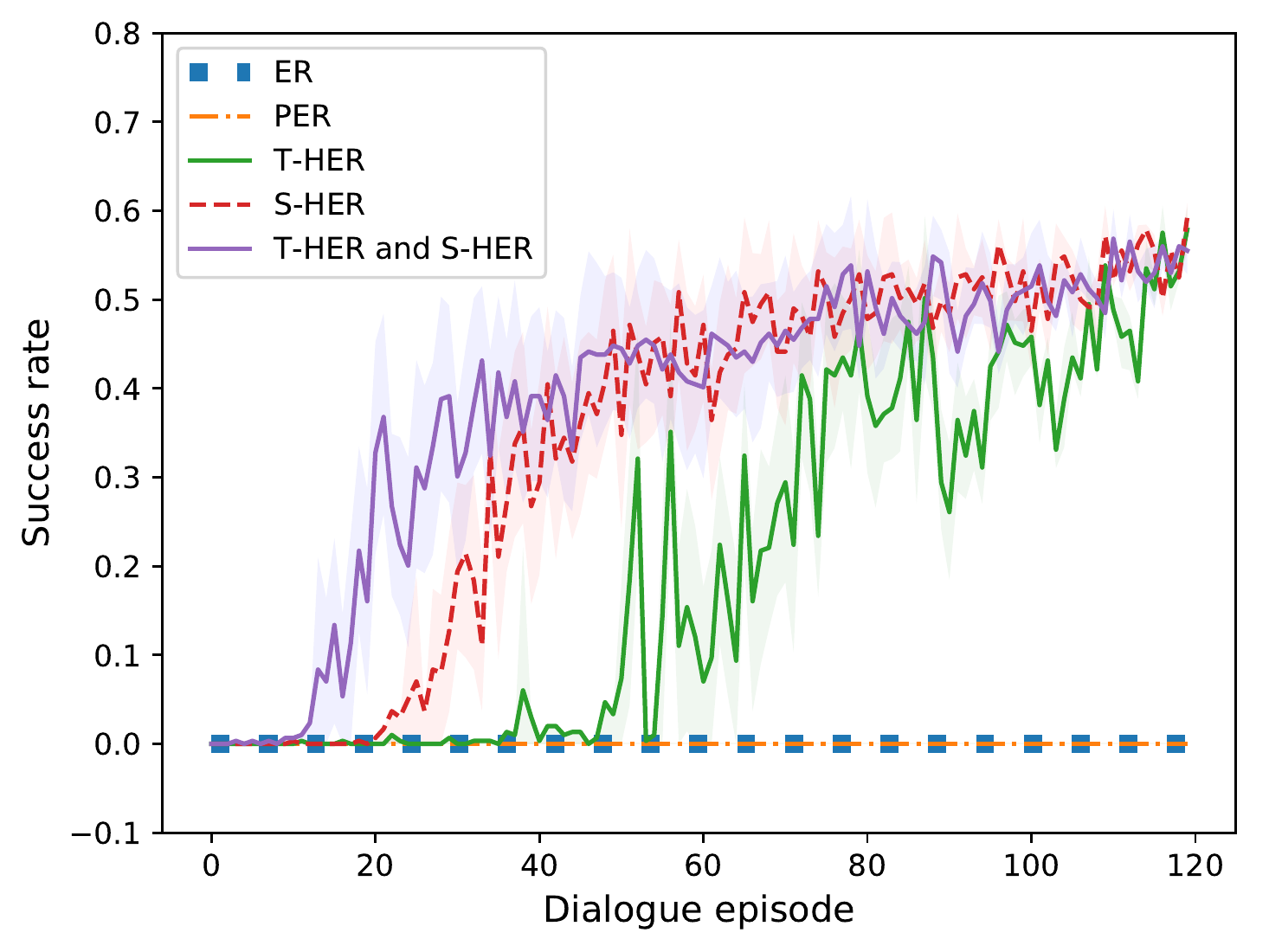}
\vspace{-1em}
\caption{In extreme situations (cold start and small experience pool), S-HER performs better than T-HER, and their combination performs the best. }
\label{fig:withoutwarmstart}
\end{figure}

Figure~\ref{fig:withoutwarmstart} shows the results. We can see that uniform experience replay (ER)~\cite{mnih2015human}, and PER~\cite{schaul2015prioritized} could not accomplish any task.
Under such challenging settings, our complex HER methods achieved $>0.5$ success rates.
In particular, S-HER outperforms T-HER, by generating significantly more successful dialogues.
Finally, combining T-HER and S-HER produced the best performance.


\section{Conclusions and Future Work}
In this work, we developed two complex hindsight experience replay (HER) methods, namely Trimming-based HER and Stitching-based HER, for dialogue data augmentation (DDA).
Our two HER methods use human-agent dialogues to generate successful, artificial dialogues that are particularly useful for learning when successful dialogues are rare (e.g., in early learning phase).
We used a realistic dialogue simulator for experiments.
Results suggest that our methods significantly increased the learning rate of a DQN-based reinforcement learner, and incorporating other experience replay methods further improved their performance.

This is the first work that applies the HER idea to the problem of dialogue policy learning.
In the future, we plan to evaluate our complex HER strategies using other dialogue simulation platform, e.g., PyDial~\cite{ultes2017pydial}, and other testing environments. Another direction is to evaluate the robustness of T-HER and S-HER by replacing DQN with other RL algorithms, such as Actor Critic~\cite{su2017sample}. Finally, we will improve this line of research by further considering the noise from language understanding.

\section*{Acknowledgements}
This work is supported in part by the National Natural Science Foundation of China under grant number U$1613216$ and Double-First Class Foundation of University of Science and Technology of Chain.

{\small
\bibliographystyle{aaai}
\bibliography{references}
}
\end{document}